\documentclass[10pt,twocolumn,letterpaper]{article}

\usepackage{cvpr}
\usepackage{times}
\usepackage{epsfig}
\usepackage{graphicx}
\usepackage{amsmath}
\usepackage{amssymb}
\usepackage[usenames]{color}

\usepackage{caption}
\usepackage{subcaption}

\usepackage{algorithm}
\usepackage{algpseudocode}

\algnewcommand{\Inputs}[1]{%
  \State \textbf{Inputs:}
  \Statex \hspace*{\algorithmicindent}\parbox[t]{.8\linewidth}{\raggedright #1}
}

\algnewcommand{\Outputs}[1]{%
  \State \textbf{Outputs:}
  \Statex \hspace*{\algorithmicindent}\parbox[t]{.8\linewidth}{\raggedright #1}
}

\algnewcommand{\Initialize}[1]{%
  \State \textbf{Initialize:}
  \Statex \hspace*{\algorithmicindent}\parbox[t]{.8\linewidth}{\raggedright #1}
}

\usepackage[breaklinks=true,bookmarks=false]{hyperref}

\cvprfinalcopy

\def\cvprPaperID{1234} 

\ifcvprfinal\fi
\begin{document}

\title{Semantic Facial Expression Editing using Autoencoded Flow}

\author{Raymond Yeh$^*$ \quad
Ziwei Liu$^{\dagger}$ \quad
Dan B Goldman$^{\ddagger}$ \quad
Aseem Agarwala$^{\ddagger}$ \\ \vspace{0.02cm}
\and
University of Illinois at Urbana-Champaign$^*$ \\
{\tt\small yeh17@illinois.edu}
\and 
The Chinese University of Hong Kong$^{\dagger}$ \\
{\tt\small lz013@ie.cuhk.edu.hk}
\and
Google Inc.$^{\ddagger}$\\
{\tt\small \{dgo,aseemaa\}@google.com}
}

\maketitle

\begin{abstract}
High-level manipulation of facial expressions in images --- such as
changing a smile to a neutral expression --- is challenging because facial
expression changes are highly non-linear, and vary depending on the
appearance of the face. We present a fully automatic approach to editing
faces that combines the advantages of flow-based face manipulation with
the more recent generative capabilities of Variational Autoencoders
(VAEs). During training, our model learns to encode the flow from one
expression to another over a low-dimensional latent space. At test
time, expression editing can be done simply using latent vector
arithmetic. We evaluate our methods on two applications: 1) single-image
facial expression editing, and 2) facial expression interpolation between
two images. We demonstrate that our method generates images of higher perceptual
quality than previous VAE and flow-based methods.

\end{abstract}

\section{Introduction}
\begin{figure}[h]
\begin{center}
\includegraphics[scale=0.45]{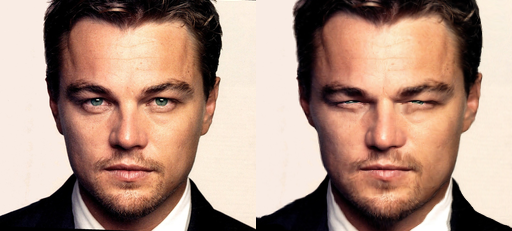}
\end{center}

\caption{Illustration of the facial expression manipulation task. \textbf{Left:} Source image.\protect\footnotemark  \textbf{Right:} Squint expression synthesized automatically by our method.}
\label{fig:intro}
\end{figure}
\footnotetext{Image obtained from \url{http://pic2.me/wallpaper/7683.html}}

The pixel-level manipulation of photographs is a common task that has received considerable attention, both in the form of commercial products and research papers. High-level, \emph{semantic} editing of images, such as turning a frown into a smile, is much less well-understood, because it requires holistic understanding of how images correspond to linguistic concepts. Recent and dramatic advances in image understanding, however, suggest that we can start performing these types of semantic image edits.  

In this paper we address the problem of high-level facial expression editing, such as changing emotions. For example, we take a neutral face photo and make it smile, or squint, or exhibit disgust. We can also exaggerate an existing expression, or perform the opposite task of making an expression subtler. Finally, we also show how to interpolate between two facial expressions, while maintaining a more realistic image than a simple cross-fade or morph would afford.

Our approach to realistic facial expression manipulation is inspired by recent facial synthesis work using convolutional neural networks (CNNs)~\cite{larsen2016autoencoding,yan2016attribute,ghodrati2015towards}. These techniques model the space of realistic faces by auto-encoding a dataset of faces down to a low-dimensional latent vector; they then generate a face image from this latent vector. The advantage of this latent space is that it is more linear than the space of general images~\cite{reed2015deep, larsen2016autoencoding}. When this space is well-formed, sampling novel latent vectors yields novel but realistic face images. Also, interpolating latent vectors yield realistic interpolated faces, and certain directions in the latent space are aligned with semantic properties, such as smiles, or beards.

These methods show impressive and exciting generalization capabilities, and can generate recognizable faces over a wide range of identities and expressions. However, close inspections of these generated images often reveal low resolution, blurriness, and broken facial features. Not surprisingly, synthesizing faces by hallucinating RGB values from scratch is very difficult. An alternative and less ambitious approach, long employed by the graphics community~\cite{Mohammed:2009:VGN,Barnes:2009:PAR}, is to generate or manipulate images by re-using parts of existing ones. While these results are often high-resolution and very realistic, the lack of a latent space prevents more high-level, semantic operations.

In this paper we explore whether the strengths of these two methods can be combined. Inspired by recent techniques that place differentiable optical flow layers within neural networks~\cite{jaderberg2015spatial,zhou2016view}, we train an autoencoder that learns to synthesize face images by flowing the pixels of existing face images. Instead of a latent space that directly encodes RGB values of face images, our latent space encodes how to flow one face image to another of the same person. We show that this approach allows us to maintain the high-resolution, sharp textures of existing images while manipulating images in a meaningful latent space.

\section{Related Work}
Deep neural network methods for image synthesis have recently produced impressive results~\cite{larsen2016autoencoding,yan2016attribute,ghodrati2015towards,denton:deep,radford2015unsupervised,pathakCVPR16context}. Many of these techniques employ an additional adversarial network to further optimize the appearance of the generated image~\cite{denton:deep,radford2015unsupervised}. The most similar paper to our work is DeepWarp~\cite{ganin:deepwarp}, which also uses neural networks to synthesize flow fields that manipulate facial expression. However, this technique is specifically limited to changing the gaze direction of eyes in an image, most typically for video-conferencing applications. 

Traditional graphics techniques for editing and synthesizing facial expressions operate by manipulating patches of existing images. The Visio-lization method~\cite{Mohammed:2009:VGN} hallucinates novel faces by compositing facial patches from a large dataset. Optical flow computed over 3D face models allow facial expression manipulation and compositing of expressions~\cite{yang2012facial,yang2012face,yang2011expression}. Beyond faces, feature correspondence has long been used to warp and morph images~\cite{Beier:1992:FIM}. We compare the results of our CNN-based method against both morphing and optical flow in Section~\ref{sec:results}.

\section{Approach}
Deep generative models, such as Variational Autoencoders (VAEs)
\cite{kingma2013auto}, have demonstrated that image generation and editing can be
done through vector arithmetic in the latent space
\cite{yan2016attribute, radford2015unsupervised, larsen2016autoencoding}.
Inspired by those models, we proposed the Flow Variational Autoencoder (FVAE)
model, which is a hybrid of VAE and the classical flow-based approach
\cite{yang2011expression, yang2012facial}. In particular, the decoder of
our FVAE learns a mapping from the latent space to the flow space, rather than pixel space.

The FVAE takes a pair of source and target images as input and outputs a per-pixel flow field and a per-pixel confidence mask. The flow field warps the
source image to be \textit{similar} to the target image, and the confidence mask
indicates how confident the model is with its flow predictions. 

In the following sections, we describe our objective functions to train a FVAE,
and how to utilize a trained FVAE for two applications: 1) single image facial
expression editing and 2) facial expression interpolation between two face images. 

\subsection{Learning Flow Variational Autoencoder}
The objective function for learning a FVAE consists of three parts: 1)
$\mathcal{L}_{reconstruct}$, which controls the pixel-wise image difference,
2) $\mathcal{L}_{prior}$, which controls the smoothness in the latent space, and
3) $\mathcal{L}_{flow}$, which controls the coherence in the flow space.
The overall objective is to minimize the following loss function,
\begin{equation}
\mathcal{L} = \mathcal{L}_{reconstruct} 
						+ \lambda_1 \mathcal{L}_{prior}
						+ \lambda_2 \mathcal{L}_{flow}
\end{equation}
where $\lambda_1$ and $\lambda_2$ are hyperparameters to be tuned. 

\subsubsection{FVAE Variational Autoencoder}
We first denote the following notations that will be used throughout the
formulation: $T$ = target image, $S$ = source image, $f$ = encoder,
$g$ = decoder, $D$ = set of $(S, T)$ tuples in the dataset, and 
$Z$ = latent vector. \\
\textbf{Encoder Network $f(T)$:} Following the VAE framework
\cite{kingma2013auto}, we assume the prior distribution follows a
multivariate Gaussian distribution. Using the reparameterization trick,
our model's encoder network, $f(T) = (\boldsymbol{\mu}(T), \boldsymbol{\sigma}(T))$,
outputs the mean $\boldsymbol{\mu}$, and
standard deviation $\boldsymbol{\sigma}$ of the approximate posterior.
To compute the latent vector $Z$ for a given $T$, we sample a random $\epsilon$
from the multivariate Gaussian distribution, $\mathcal{N}$.
\begin{equation}
Z(T) = \boldsymbol{\mu}(T) + \boldsymbol{\sigma}(T) \odot \boldsymbol{\epsilon}
\end{equation}
where, $\boldsymbol{\epsilon} \sim \mathcal{N}(0, \mathbf{I})$. \\
As proposed and derived in \cite{kingma2013auto}, $\mathcal{L}_{prior}$ has the
following form,
\begin{equation}
\mathcal{L}_{prior} = -\sum_{T \in D}\sum_{k}
\bigg{(}1+\log(\sigma_k^2(T))-\mu_k^2(T)-(\sigma_k^2(T))\bigg{)},
\end{equation}
where $k$ indexes over the dimensions of the latent space. \\
\textbf{Decoder Network $g(S,Z(T))$:} The proposed FVAE can be viewed as a
constrained version of VAE, as the output of FVAE can only contain pixels flowed
from the source image.
\begin{equation} \label{eq:l_reconstruct}
\mathcal{L}_{reconstruct} = \sum\limits_{(S, T) \in D} || T - g(S,Z(T))||_2^2
\end{equation}
subject to $g_j(S,Z(T)) \in \{S\} \quad \forall j$, where $j$ denotes the pixel
location.

In order to satisfy the constraint the network computes a dense flow field,
$F$, as an intermediate state. Denote $F(j) \triangleq (F(j)_x, F(j)_y)$, which
specifies the pixel location to sample from $S$ to reconstruct the 
$j^{th}$ pixel of the output, $g_j(S,Z(T))$. To train the network from
end-to-end, we relax the constraint in Eq.\ref{eq:l_reconstruct}, such that
$g_j(S,T)$ is in the set of bilinear interpolated values from $\{S\}$
~\cite{zhou2016view}. The decoder network, $g$, has the form

\begin{align}
&g_j(S,Z(T)) = \notag \\ &\sum_{k \in N(F(j))} S_k (1-|F(j)_x- x_k|)(1-|F(j)_y-y_k|),
\end{align}
where $(F(j)_x, F(j)_y)$ denotes the absolute coordinates to sample from the
source image, $N(F(j))$ denotes the set of the 4 neighbors of $F(j)$, $S_k$ 
denotes the $k^{th}$ pixel of the source image, and $(x_k,y_k)$ denotes the pixel
absolute coordinates of the $k^{th}$ pixel.

This bilinear sampling mechanism was introduced by Jaderberg \etal as
differentiable image sampling with a bilinear kernel; Error backpropagation
can be done efficiently through this mechanism \cite{jaderberg2015spatial}. 

\subsubsection{Flow Coherence Regularization}
Our choice of the flow coherence regularizer, $\mathcal{L}_{flow}$, is based
on the intuition that flow directions should be spatially coherent except at
boundaries. We design a pair-wise regularization term as
follows: 
\begin{align}\label{eq:flow_coherence}
&\mathcal{L}_{flow} =\\ \notag & 
\sum\limits_{\mathbf{i}}\sum\limits_{\mathbf{j} \in N(\mathbf{i})} 
||F_\mathbf{i}-F_\mathbf{j}||_2^2
\exp(-(\alpha||T_{\mathbf{i}}-T_{\mathbf{j}}||_2^2
+ ||\mathbf{i}-\mathbf{j}||_2^2))
\end{align}\\
where, $\mathbf{i}, \mathbf{j}$ are 2D-coordinates, $F_{\mathbf{i}}$ is the flow
vector at $\mathbf{i}$, $T_{\mathbf{i}}$ is a RGB color vector of the target
image at $\mathbf{i}$, $\alpha$ is a weighting hyper-parameter, and
$N(\mathbf{i})$ is the set of neighboring pixels of $\mathbf{i}$.
In our experiment we use a  $7\times7$ neighborhood centered at $\mathbf{i}$. 

Eq.\ref{eq:flow_coherence} implies that the flow is encouraged to be similar
for nearby pixel coordinates, especially when RGB values are similar in the target image. Similar
smoothness formulations have been used in the context of semantic image segmentation
\cite{zheng2015conditional, schwing2015fully, chen2014semantic, liu2015dpn}. 

\subsection{Learning the Confidence Mask}
From a single source image, it may not be possible to form a good reconstruction
of the target image. For example, it will be very difficult to reconstruct a
smile expression given a neutral expression, as the teeth region simply does not
exist in the source image. Ideally, we would like to have a mechanism to
identify the regions where the network will perform poorly given a particular pair
of source and target image. With such a mechanism, we would be able combine
multiple source images to generate one output image.

In order to identify such regions, we trained another network to predict a
per-pixel soft confidence mask $M$ where $M_j$ predicts the confidence of the
generated output $g_j(S,Z(T))$. This confidence-learning problem can be formulated
as a binary classification task, where the label is 1 if the target pixel is
in the source image, and 0 otherwise. However, getting such labels would require
searching each pixel pair in $S$ and $T$ for all $(S, T)$ pairs, and is
not scalable. Instead we choose to approximate the true label, $y$,
using the $\ell_2$-norm, as follows:

Given a trained encoder $f$ and decoder $g$,
\begin{equation}
y = \exp\bigg{(}-\frac{||T-g(S,Z(T))||_2^2}{\beta}\bigg{)}
\end{equation}
, where $\beta$ is a normalization constant, such that $y\in [0,1]$. 

The confidence mask network takes $Z(T)$ and $S$ as inputs (\emph{i.e}\onedot, $M(S,Z(T))$),
as at test-time the target image is not available. With the approximate true
label $y$, the objective function for learning the confidence mask is the sum of
cross entropy losses over all pixel locations. To train the network we
minimize the following loss function: 

\begin{equation}
\mathcal{L}_{mask} = -\sum\limits_{(S,T) \in D}\sum\limits_{\mathbf{i}}
y_{\mathbf{i}} \log(M_{\mathbf{i}}) + (1-y_{\mathbf{i}}) \log(1-M_{\mathbf{i}}) 
\end{equation}

\section{Applications}
With a trained FVAE model, we proposed two applications 1) expression editing on
a single source image, and 2) expression interpolation from a pair of source 
images.

\subsection{Expression Editing Using Latent Vector}
\begin{figure}[h]
\begin{center}
\includegraphics[scale=0.6]{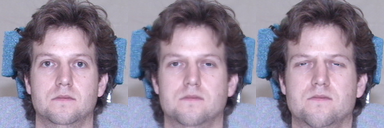}
\end{center}
\caption{Traversal along the ``squint direction''. The left-most image is the
original image; the other two images are synthesized squint expressions at
different ``intensities'' of squint.}
\label{fig:squint_traverse}
\end{figure}

For this task, the goal is to transform the expression of a given face image to
a desired target expression; see Figure \ref{fig:squint_traverse} for
illustration. The task is an analogy making problem, A:B::C:D means A is to B as
C is to D \cite{reed2015deep} (i.e., generate a D such that the
relationship between C and D is similar to that of A and B).

Latent representations from deep models have been shown to capture semantic and
visual relationships, and analogy can be made through vector arithmetic; if
A:B::C:D then $(A-B) \approx (C-D)$. For example, in language models, the vector
representations of ``King" - ``Man" + ``Woman" is a vector similar to ``Queen."
\cite{mikolov2013distributed, mikolov2013linguistic}. Similarly, visual
analogy making has been explored in
\cite{reed2015deep, radford2015unsupervised}.

Algorithm \ref{alg:exp_edit} details the method for performing expression
editing using the proposed FVAE. The inputs are: 1) the trained FVAE network,
$(f,g)$, 2) training dataset face images, $I$, and expression labels, $y$, 3)
a source image for editing, $S$, 4) a target expression label, $y_{target}$, 5)
the source expression label, $y_{source}$, and 6) the number of steps, $K$.

\begin{algorithm}
  \caption{Expression Edit}
  \begin{algorithmic}[1]
    \Inputs{$f,g$, $D=\{(I,y)\}$, $S$, $y_{target}$, $y_{source}$, $K$}
    \Initialize{$Z_{target} = \frac{1}{|D|}\sum\limits_{D} Z(I) 
                              \cdot \mathbf{1}[y=y_{target}]$}
    \Initialize{$Z_{source} = \frac{1}{|D|}\sum\limits_{D} Z(I)
                              \cdot \mathbf{1}[y=y_{source}]$}
    \For{each node $i = 1$ to $K$}
    \State $Z_{edit} = \frac{i \cdot (Z_{target} - Z_{source})}{K} + Z(S)$
    \Outputs{$g(S,Z_{edit})$}
    \EndFor
  \end{algorithmic}
  \label{alg:exp_edit}
\end{algorithm}

Algorithm \ref{alg:exp_edit} computes a latent \textit{expression} direction by
taking the difference of the averaged latent vectors for training images with 
$y_{target}$ expression label, and training images with 
$y_{source}$ expression label. Next, the source image's latent vector,
$Z_{source}$, is interpolated linearly along the \textit{expression} direction.

\subsection{Expression Interpolation Using Latent Vector}
The generative power of VAE models is often demonstrated by performing 
linear interpolation between two given locations in the latent space. For this
task, the goal is to generate all the intermediate ``levels'' of expression
between two given source images; see Figure \ref{fig:image_interp} for
illustration. Algorithm \ref{alg:exp_interp} details the
method for expression interpolation. The inputs are: 1) the two source images
$S_1$ and $S_2$, 2) a trained FVAE $(f,g)$, 3) a trained mask model, $M$, and
4) the number of steps, $K$. 

In order to utilize both source images in the linear interpolation, we use the
normalized confidence mask to blend the interpolated outputs from each source image
together. This is useful when certain facial parts are only in one of the source
image, (\emph{e.g}\onedot, teeth when interpolating from neutral to smile). 

\begin{algorithm}
  \caption{Expression Interpolate}
    \begin{algorithmic}[1]
    \Inputs{$f,g$, $S_1$, $S_2$, $M$, $K$}
    \Initialize{$Z_1 = Z(S_1)$}
    \Initialize{$Z_2 = Z(S_2)$}
    \For{each node $i = 1$ to $K$}
    \State $Z_{edit} = \frac{i \cdot (Z_2 - Z_1)}{K} + Z_1$
    \State $O_1 = g(S_1,Z_{edit})$
    \State $O_2 = g(S_2,Z_{edit})$
    \State $M_1 = M(S_1,Z_{edit})$
    \State $M_2 = M(S_2,Z_{edit})$
    \Outputs{$\frac{\exp(M_1)}{\exp(M_1+M_2)} \odot O_1 +
              \frac{\exp(M_2)}{exp(M_1+M_2)} \odot O_2$}
  \EndFor
  \end{algorithmic}
  \label{alg:exp_interp}        
\end{algorithm}

\subsection{Flow Upsampling for High Resolution Output}
Observe that one limitation of using a deep model is that the output dimensions
are fixed. A model trained to generate $128 \times 128$ images cannot output
$512 \times 512$ images. To get higher resolution output, typically, one will
have to upsample in the pixel domain; this often introduces blurriness. Here we
proposed to perform upsampling in the \textit{flow domain} instead. We assume that
high-resolution source images are available. Applying the upsampled flow to
the high-resolution source images tends to produce sharper images, see Figure
~\ref{fig:flow_upsample} for comparison.

\section{Implementation Details}
\begin{figure*}[t]
  \centering
  \includegraphics[width=0.95\textwidth]{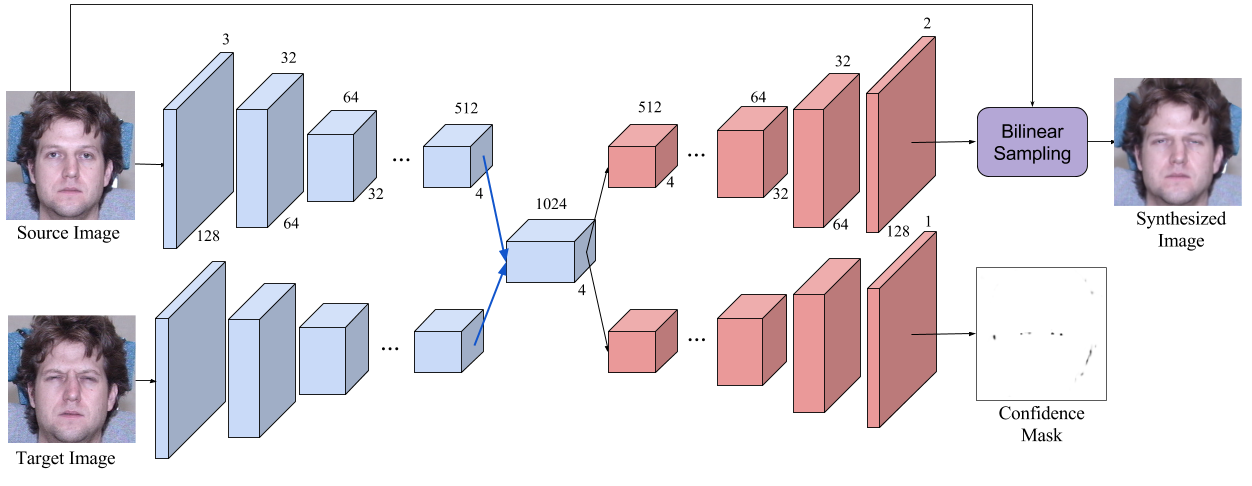}
  \caption{Detailed FVAE network architecture. The blue blocks denote
  convolution layers, and the red blocks denote convolution upsample layers.
  The blue arrows indicate concatenation. The network design follows the
  encoder-decoder structure in \cite{radford2015unsupervised}, where the latent
  space, $Z$, is the central  $4 \times 4 \times 1024$ block.}
  \label{fig:net}
\end{figure*}

In this section, we describe the data processing and training implementation
details. 
\subsection{Dataset}
To evaluate our proposed method, we conducted experiments on the CMU MULTI-PIE
Dataset \cite{gross2010multi}, which contains the identity labels to train a
FVAE, as well as the expression labels necessary at test time.

\subsection{Data Preparation}
We use all the examples with camera view $05\_1$, and illumination $06$ from
the MULTI-PIE dataset. The training and testing set is divided based on
identity, with 80\% and 20\% random split. We aligned the faces using landmarks
detected from \cite{kazemi2014one}, then cropped and resize the images to
dimension of $128 \times 128$. To prepare the training example pairs of $(S,T)$,
we created all pairs of expression per identity. For each source and
target image pair we apply a color transform \cite{shirley2001color} to adjust
the color of the source image to better match the target images; this makes the
learning easier, as $\ell_2$-norm loss is sensitive to lighting and color changes.
We also applied data augmentation by randomly flipping images horizontally.
Lastly, we normalized the pixel values into range of [-1,1] for the ease of
training \cite{radford2015unsupervised}. 

\subsection{Model Architecture}
Our model architecture follows a similar design spirit as
\cite{radford2015unsupervised}. For the encoder, convolution layers channel
sizes double as the spatial size halves. For the decoder, convolution layers 
channel sizes halves and spatial size doubles. We used nearest neighbor 
interpolation for upsampling in the decoder. Figure \ref{fig:net}, shows the 
exact architecture used. The convolution layers' filter sizes are $3 \times 3$,
with the exception of the first and final layers, which are $5 \times 5$ filters.
Each convolution layer uses $relu$ except for the final layer outputting the
flow which uses $tanh$, as our network outputs normalized coordinates.
Additionally, all convolution layer uses batch-normalization \cite{ioffe2015batch},
except at the final two layers before the output. All the weights are initialized
from a truncated Gaussian distribution with standard deviation of 0.01. 

\subsection{Training}
We train the network using the Adam optimizer \cite{kingma2015adam}, with
learning rate of $0.0003$, $\beta_1 = 0.9$, $\beta_2= 0.999$ and mini-batch size
of $64$. We use gradient clipping by value with threshold of $\pm 5$. 
Empirically, we chose $\lambda_1 = 0.003$ and $\lambda_2 = 0.001$.
The training and validation loss are monitored to use early stopping to prevent
overfitting. The models are implemented using
TensorFlow~\cite{tensorflow2015-whitepaper}.

\section{Evaluations}
\label{sec:results}
In this section, we provide comparison experiments with qualitative and 
quantitative results for the single-image editing task and image interpolation
task. 

\subsection{Facial Expression Editing}
\begin{figure*}[t]
\begin{center}
\includegraphics[scale=0.54]
{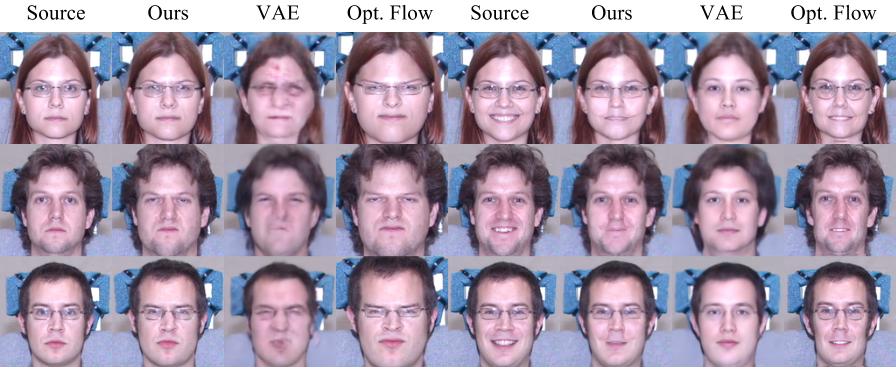}
\end{center}
\caption{Comparison of synthesized expressions. \textbf{Left block:} Synthesis from a neutral expression source image to expression of disgust. \textbf{Right block:} Synthesis from a smile expression source image to a neutral expression. Our method demonstrates better low-level image quality and more realistic expressions compared to the baselines.}
\label{fig:pie_exp3}
\end{figure*}

For the facial expression editing task, we compare our FVAE model with
both the VAE\cite{kingma2013auto} using
\cite{radford2015unsupervised}'s architecture, and an optical flow
method \cite{revaud2015epicflow}. Figure \ref{fig:pie_exp3} compares
the expression editing results on examples from the test set. Note
that for all results in our paper the person shown in the result is
part of hold-out data; no images of that person were in the training
set.  For VAE, we train a model with the same architecture as our
FVAE, except the last layer is replaced to predict the RGB pixel
values directly. Then, we use Algorithm \ref{alg:exp_edit} to perform
the editing. For optical flow, we use the average flow extracted from
all identities in the training set. It can be observed that our
method generates superior image quality compare to the other
methods. VAE results in blurry images, and the optical flow method cannot
get a flow conditioned on the source image, thus the flow is not
\textit{tailored} to the source image and creates artifacts.

For quantitative evaluation, the PSNR for each method is shown in Table
\ref{table:psnr_exp_edit}. In order to measure PSNR, we use the ground truth
target image's latent vector for FVAE and VAE. For optical flow,
using a state-of-the-art method~\cite{revaud2015epicflow}, we compute the
flow field using the ground truth target image, and warp the source image
accordingly.  

It is not surprising that a VAE obtains the highest PSNR, as it is directly
generating pixels to minimize the $\ell_2$-norm loss. 
However, judging by the perceptual quality, our FVAE significantly outperforms 
the VAE; VAE generates overly smooth images with poor low-level image quality
(e.g., hair and skin textures).
This just demonstrates that PSNR is not an ideal quantitative measurement for
this task. 

\begin{table}[h]
\begin{center}
\begin{tabular}{|c|c|}
\hline
Method & PSNR \\ 
\hline
Ours & 18.14 \\
\hline
Ours w/ Confidence& \textit{18.35} \\
\hline 
VAE \cite{kingma2013auto, radford2015unsupervised}&  \textbf{25.41} \\
\hline
Epic Flow \cite{revaud2015epicflow}& 15.76 \\
\hline
\end{tabular}
\end{center}
\caption{PSNR comparison between different methods. Higher is better.}
\label{table:psnr_exp_edit}
\end{table}

\begin{figure}[h]
\begin{center}
\includegraphics[scale=0.3]{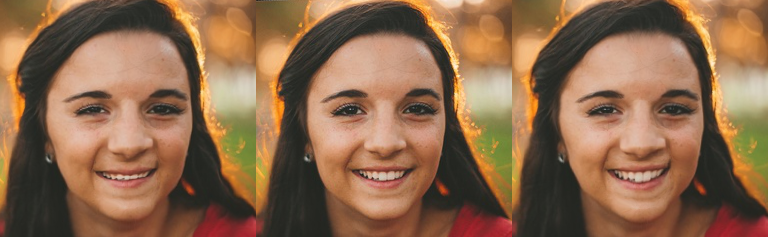}
\includegraphics[scale=0.3]{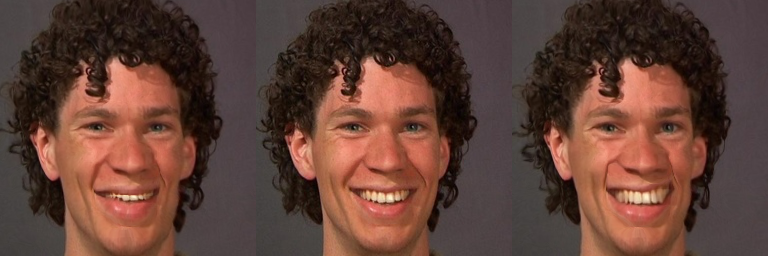}
\includegraphics[scale=0.3]{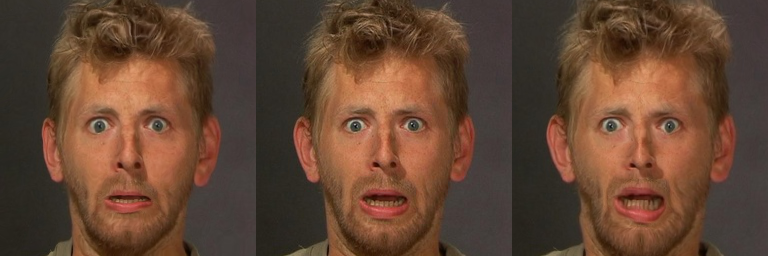}
\end{center}
\caption{Magnify and suppress the facial expression. 
				 \textbf{Left}: Synthesized suppressed expression.
				 \textbf{Center}: Original image.\protect\footnotemark
				 \textbf{Right}: Synthesized magnified image.}
\label{fig:magnify_supress}
\end{figure}

Next, we demonstrate that the proposed FVAE method can be transferred to 
\textit{out-of-dataset} samples and expressions. Figure \ref{fig:magnify_supress} 
demonstrates that our method can be used to edit web images to suppress or
magnify the facial expressions. The third row in Figure 
\ref{fig:magnify_supress}, is the expression of fear, which is not in the 
dataset; however, the editing can be still be done by approximating 
$Z_{source} = Z(S)$, in Algorithm \ref{alg:exp_edit}. Note that for web-images 
to generate reasonable quality images, the background needs to be clean and 
the face region should not contain significant lighting variations, such as shadows. 

\footnotetext{Images from \cite{yang2012facial} and\\ \url{http://www.soalinejackphotography.com}}

\subsection{Facial Expression Interpolation}
\begin{figure*}[t]
\begin{center}

\includegraphics[scale=1.0]{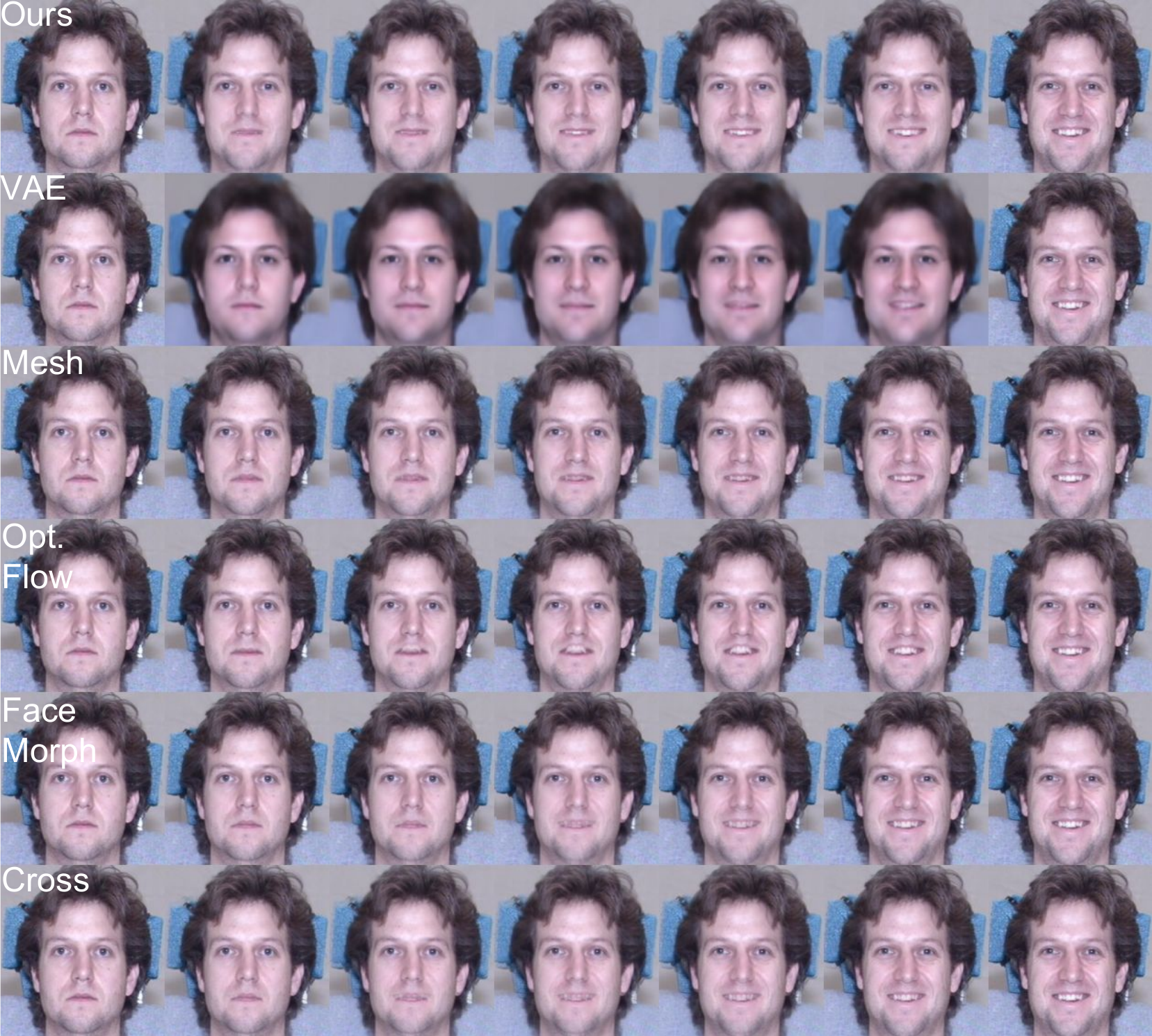}

\end{center}
\caption{Image interpolation comparison. Images at both ends are the source images,
and all images in between are synthesized.
\textbf{Row 1}: Our method.
\textbf{Row 2}: VAE.
\textbf{Row 3}: Mesh Morphing~\cite{wolberg1996recent}.
\textbf{Row 4}: Optical Flow~\cite{liu2009beyond}.
\textbf{Row 5}: Face Morphing~\cite{yang2012face}.
\textbf{Row 6}: Cross Fading. Our results are perceptually better than all the baselines. The most challenging area is the mouth and teeth, because of their complicated motion and occlusions/dis-occlusions.
}
\label{fig:image_interp}
\end{figure*}

For the facial expression interpolation task, we compared our method
against five different methods: VAE, cross-fading, mesh morphing
\cite{wolberg1996recent}, optical flow \cite{liu2009beyond}, and face
morphing \cite{yang2012face}.  Figure \ref{fig:image_interp} shows the
qualitative comparison between the methods on examples from the test
set. It can be observed that our method produces the sharpest and most
realistic image out of all methods. The confidence mask successfully
handles the teeth regions, whereas morphing-based
methods~\cite{wolberg1996recent,yang2012face} and optical
flow~\cite{liu2009beyond} generate blurry results around the mouth
regions. Classical mesh morphing produces the second-best results, but
still has trouble with the teeth. Moreover, our method provides a more
coherent and natural transition between the expression levels.

\subsection{Perception User Study}
We perform a perception user study to compare the photorealism of real images,
images generated from our proposed method, VAE, and optical flow. We perform
an experiment asking Amazon Mechanical Turk's workers to label images as real or
fake. The experiment is conducted on 25 participants; each is asked to label 10
examples generated from each of the methods, and 10 real images.
the images are provided in random order on a single web page. To guarantee the quality
of the results, users performing below 50\% accuracy on the real images are dropped
from the experiment. 
Additionally, some users skipped over certain images;
we also do not count the skipped images in our result. The percentage of images
labeled as real for each of the method is 

reported in Table~\ref{table:user_result}.
As can be seen, our method outperforms VAE and optical flow in perceptual quality. 

\begin{table}[h]
\begin{center}
\begin{tabular}{|c|c|}
\hline
Method & \% labeled real \\ 
\hline
Real & 89.7\% \\
\hline
Ours & 59.4\% \\
\hline
VAE & 35.6\% \\
\hline
Opt. Flow & 41.6\% \\
\hline
\end{tabular}
\end{center}
\caption{Photorealism comparison between different methods. Higher is better.}
\label{table:user_result}
\end{table}

\subsection{Effectiveness of Flow Upsampling}
\begin{figure}[t]
\begin{center}
\includegraphics[scale=0.23]{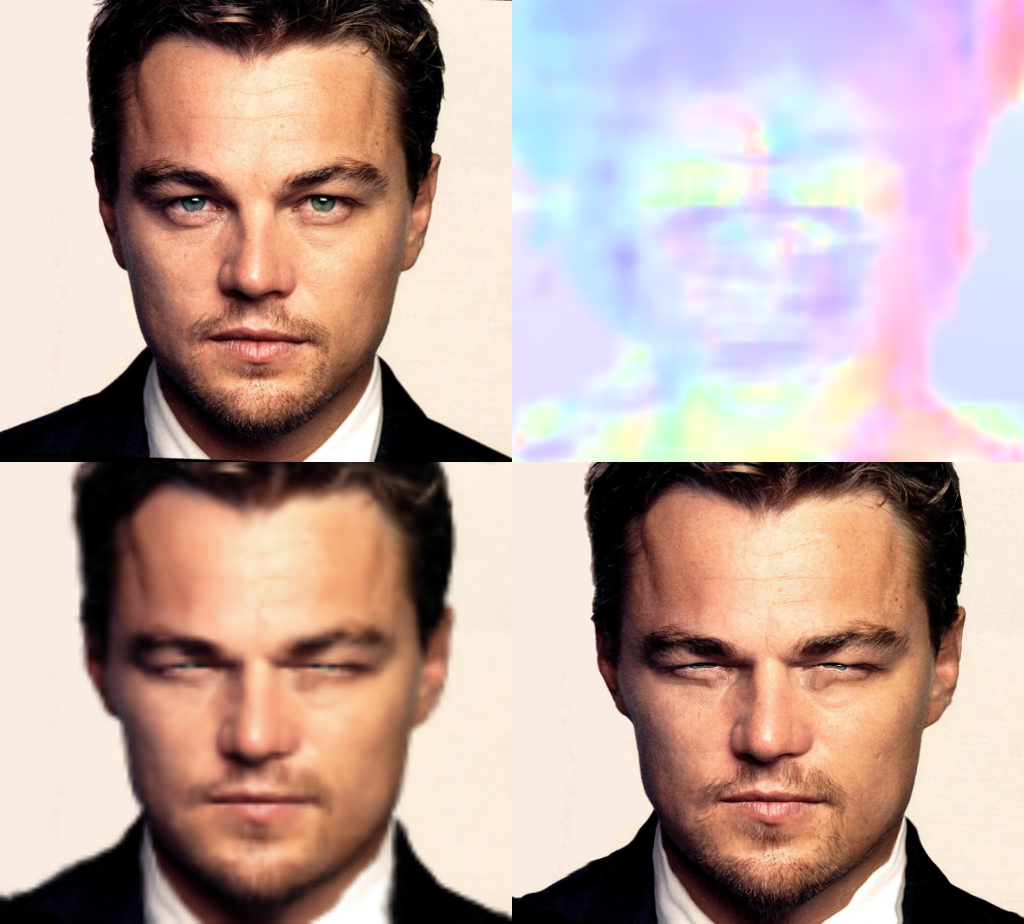}
\end{center}

\caption{Upsampling comparison.
				 \textbf{Top left}: High-res source image.
				 \textbf{Top right}: Flow visualization.
				 \textbf{Bottom left}: Pixel upsampled result. 
				 \textbf{Bottom right}: Flow upsample result.}
\label{fig:flow_upsample}
\end{figure}

Figure \ref{fig:flow_upsample} shows the comparison between upsampling in the
pixel domain versus in the flow domain. The source images is of dimension
$512 \times 512$. In order to use our trained network of dimension
$128 \times 128$, we resize the source image to $128 \times 128$. For
comparison, we upsample the output images back to $512 \times 512$ in the pixel
domain. For flow based upsampling, we upsample the flow back to $512 \times 512$
using bilinear interpolation. Next, we pass the upsampled flow through the
bilinear sampling mechanism with the high-resolution source image to synthesize
the output. As can be seen, upsampling in the pixel domain leads to blurrier
results compared to the flow upsampling method. The flow upsampling method preserves
more of the finer details, (\emph{e.g}\onedot, edges, and facial hair texture), from the
source image. We are able to upsample the flow by four times without noticeable
distortions in the output image.

\section{Conclusion}
We have presented an approach to the realistic manipulation of facial
expressions that combines the strengths of deep models and flow-based
image editing approachs. We demonstrate two applications: 1)
high-level facial expression editing from a single source image, and
2) high-level facial expression interpolation from two source
images. The proposed method successfully outperforms the baseline
methods on the MULTI-PIE dataset. Our method generates higher
resolution images, with realistic image quality and more natural
expression changes. We further demonstrate that the learned
transformation can be generalized to \textit{out-of-dataset} samples
with significantly different image statistics, while deep models
hallucinating RGB values directly fail to generalize.

Despite outperforming the baseline methods, our method still faces some
challenges: 
\begin{enumerate}
\item Our method is limited to frontal pose faces with small rotations.
Although ~\cite{zhou2016view} demonstrates 3D rotation on objects is plausible,
the task becomes more difficult when the training data does not contain a
fine scales of rotated images. For example, the MULTI-PIE dataset contains
45-degree jumps between the rotation angles, and this lead to difficulties
during training. However, we believe that our method can properly handle
rotation given a more suitable dataset.

\item Training data needs to be very controlled. Our method requires face
images taken under similar lighting conditions and backgrounds. The main reason
is that $\ell_2$-norm loss is sensitive to RGB color changes. Without controlled
data, the model will focus on learning the lighting or background difference,
and less on the expression. One possible solution is to utilize
face and lighting information to design a lighting and background invariant loss
function.
\end{enumerate}

We leave these challenges for future work.

{\small
\bibliographystyle{ieee}
\bibliography{egbib}
}

\end{document}